\documentclass{article}
\PassOptionsToPackage{numbers, compress}{natbib}
 \usepackage[preprint]{neurips_2026}

\usepackage[utf8]{inputenc} 
\usepackage[T1]{fontenc}    
\usepackage{hyperref}       
\usepackage{url}            
\usepackage{booktabs}       
\usepackage{amsfonts}       
\usepackage{nicefrac}       
\usepackage{microtype}      
\usepackage{xcolor}         
\usepackage{graphicx}
\usepackage{multirow}
\usepackage{makecell}
\usepackage[table]{xcolor}
\usepackage{amsmath}
\usepackage{caption}
\usepackage{wrapfig}
\usepackage{enumitem}
\usepackage{placeins}
\title{\textit{Retrieve, Integrate, and Synthesize}: Spatial-Semantic Grounded Latent Visual Reasoning}

\author{
    Jin Cui\textsuperscript{$*1$}, 
    Xinyue Long\textsuperscript{$*2$}, 
    Xunyong Zhang\textsuperscript{$1$}, 
    Yadong Zhang\textsuperscript{$1$}, \\
    \textbf{Chuanchang Su}\textsuperscript{$1$},
    \textbf{Jingye Gan}\textsuperscript{$1$}, 
    \textbf{Boran Zhao}\textsuperscript{$\dagger2$},
    \textbf{Pengju Ren}\textsuperscript{$1$} \\[1mm]
    \textsuperscript{$1$}State Key Laboratory of Human-Machine Hybrid Augmented Intelligence,\\
    and Institute of Artificial Intelligence and Robotics, Xi'an Jiaotong University\\
    \textsuperscript{$2$}School of Software Engineering, \\
    State Key Laboratory of Human-Machine Hybrid Augmented Intelligence,\\
    Institute of Artificial Intelligence and Robotics, Xi'an Jiaotong University\\  
    {\tt\small andycui@stu.xjtu.edu.cn, \{boranzhao, pengjuren\}@xjtu.edu.cn}
}

\begin{document}

\maketitle{
    \renewcommand{\thefootnote}{\fnsymbol{footnote}}
    \footnotetext[1]{Equal contribution.} \footnotetext[2]{Corresponding author.}  
}

\begin{abstract}

Multimodal Large Language Models (MLLMs) have made remarkable progress on vision-language reasoning, yet most methods still compress visual evidence into discrete textual thoughts, creating an information bottleneck for fine-grained perception. Recent latent visual reasoning methods attempt to reason in continuous hidden states, but we find that they suffer from insufficient manifold compatibility: latent trajectories drift away from pretrained reasoning circuits, collapse into instance-agnostic patterns, and are often bypassed during answer generation. To address these issues, we propose \textit{\textbf{RIS} (\textbf{R}etrieve, \textbf{I}ntegrate, and \textbf{S}ynthesize)}, a spatial-semantic grounded framework that develops latent reasoning as a compatible extension of pretrained MLLM computation. We first construct a step-wise grounded reasoning dataset with bounding boxes and region-specific semantic descriptions. Built on this supervision, \textbf{\textit{RIS}} anchors latent tokens to both spatial and semantic evidence, enforces their causal role through a progressive attention bottleneck, and introduces short language transition tokens to bridge synthesized latent states back to vocabulary-aligned decoding. Experiments on V$^\ast$, HRBench4K, HRBench8K, MMVP, and BLINK show consistent improvements over closed/open-source and latent reasoning baselines. Further analyses demonstrate that \textbf{\textit{RIS}} learns diverse, interpretable, and progressively integrated latent trajectories, offering a practical path toward faithful internal visual reasoning in MLLMs.
\end{abstract}

\section{Introduction}


Multimodal Large Language Models (MLLMs) have achieved remarkable success across diverse vision-language tasks, largely due to Chain-of-Thought (CoT) reasoning\citep{wei2022chain,kojima2022large}. However, these models still treat visual information as static preconditions, converting continuous visual features into discrete textual tokens and reasoning only within the textual domain\citep{li2022blip}. This creates an inherent bottleneck: \textit{fine-grained visual evidence must be compressed into language tokens before it can participate in reasoning}. Recent ``thinking with images''\citep{su2025thinking} methods alleviate this issue by injecting visual evidence through external tools\citep{yang2023mm,liu2024llava,wang2025jigsaw} or programmatic operations\citep{gupta2023visual,suris2023vipergpt,hu2024visual}, but their flexibility is limited by predefined tool interfaces and external execution. This necessitates a more unified solution to move intermediate visual reasoning inside the model, allowing it to manipulate question-relevant visual evidence directly in continuous hidden representations.

Latent visual reasoning offers a promising path toward this goal. Unlike text-based CoT, latent states provide an expressive workspace where visual patterns and abstract concepts can be represented without being discretized into language\citep{su2025thinking,yu2026latent}. Yet this freedom also introduces a fundamental tension. Since the model's reasoning behavior and decoding interface are largely shaped by language pretraining, effective latent visual reasoning must not only exploit the expressive capacity of a latent visual manifold $\mathcal{M}_{vis}$, but also remain compatible with the vocabulary-aligned manifold $\mathcal{M}_{vocab}$ where pretrained reasoning circuits and language-grounded decoding are organized. Existing methods such as LVR\citep{li2025latent} and Monet\citep{wang2025monet} take important steps by reconstructing visual tokens from latent states or generating continuous embeddings as intermediate visual thoughts, but they do not fully resolve this compatibility problem.

In this work, we first analyze \textit{why existing latent visual reasoning methods remain ineffective despite forming distinct latent visual representations}. Recent causal mediation study\citep{li2026imagination} reveals pronounced \emph{Input--Latent} and \emph{Latent--Answer} disconnects, where latent tokens are weakly grounded in visual inputs and exert limited influence on final predictions. Our empirical analysis further shows that these failures are closely tied to manifold divergence. Specifically, (1) weakly supervised hidden states may drift away from the pretrained vocabulary-aligned manifold and tend to collapse into highly similar, instance-agnostic trajectories; (2) answer tokens usually bypass latent tokens and rely directly on the input image and question; (3) the model must converge high-entropy latent visual states abruptly to low-entropy answer tokens, which can induce representation mismatch during language decoding.

To address these challenges, we propose \textit{\textbf{RIS} (\textbf{R}etrieve, \textbf{I}ntegrate, and \textbf{S}ynthesize)}, a grounded latent visual reasoning framework that develops latent space as a compatible extension of pretrained reasoning circuits rather than a detached visual manifold. To support training, we first construct a step-wise grounded visual reasoning dataset with $96k$ samples in which each reasoning step is paired with bounding-box spatial supervision and a region-specific semantic description. Built on this spatial-semantic supervision, \textbf{\textit{RIS}} structures latent tokens as directed visual evidence retrieval states: bounding-box supervision anchors \textit{where to look}, semantic alignment specifies \textit{what is seen}, and a progressive attention mask forces task-relevant evidence to flow through latent tokens instead of being bypassed during answer generation. Slots beyond the annotated reasoning steps are optimized solely through the final-answer objective, endowing them with the emergent ability to integrate and synthesize evidence retrieved by grounded slots. Finally, we demonstrate that generating a slightly elaborated answer between latent reasoning and final option-level answer acts as manifold transition tokens since it gradually reduces the entropy of reasoning paths from latent states to low-entropy answer tokens rather than abrupt degradation, while providing dense supervision during training.

We evaluate \textbf{\textit{RIS}} on five challenging visual reasoning benchmarks. \textbf{\textit{RIS}} consistently outperforms strong baselines, with particularly clear gains on tasks requiring localization, structured visual search, and multi-step perceptual reasoning. Further analyses show that \textbf{\textit{RIS}} produces more diverse, interpretable, and task-dependent latent trajectories. Our contributions are summarized as follows:

\begin{itemize}[leftmargin=*, label=$\star$, labelsep=0.5em]
    \item We provide a systematic analysis of latent visual reasoning in MLLMs, identifying the interaction between vocabulary-aligned manifold $\mathcal{M}_{vocab}$ and latent visual manifold $\mathcal{M}_{vis}$, and revealing manifold divergence, latent trajectory collapse, and answer bypassing as key obstacles.
    \item We construct an $96k$-sample \textbf{\textit{Grounded Latent Supervision Dataset (GLSD)}} and propose \textbf{\textit{RIS}}, a spatial-semantic grounded latent reasoning framework that structures latent tokens to retrieve task-relevant visual evidence while developing latent space as a compatible extension of pretrained reasoning circuits rather than a detached visual manifold.
    \item We demonstrate consistent improvements across visual reasoning benchmarks, especially on localization and multi-step visual reasoning tasks, and further show that \textbf{\textit{RIS}} learns diverse, interpretable, and progressively integrated latent reasoning trajectories with state-of-the-art performance.

\end{itemize}

\section{Related Work}

\vspace{-2mm}

\noindent \textbf{From Static Perception to Internal Visual Imagination.}
Most current MLLMs adopt text-space CoT reasoning to solve complex visual tasks, treating visual inputs as static premises for language-based inference\citep{zhang2023multimodal,wang2025multimodal}. Although effective, such methods reason through discrete text tokens, which provide an indirect and lossy representation for fine-grained visual understanding. Recent \emph{Thinking with Images}~\citep{su2025thinking} methods alleviate this limitation by using external visual tools to manipulate and inject intermediate visual evidence\citep{yang2023mm}. However, their effectiveness is constrained by the availability, design, and granularity of predefined tools. This motivates internal visual reasoning, where models reason over
visual evidence in continuous latent states rather than translating it into text or pixels.

\noindent \textbf{Latent Reasoning.} Recent studies have explored continuous latent spaces as an alternative to discrete token-level reasoning. Representative approaches include utilizing recursive hidden states for breadth-first search\citep{hao2024training}, self-distillation of explicit reasoning traces\citep{shen2025codi}, and implicit reasoning via superposed latent chains\citep{deng2026llmlatentreasoningchain}. While these methods enhance reasoning efficiency, they remain constrained in textual space. Extending them to MLLMs is non-trivial: representing visual evidence by vocabulary embeddings or weakly supervised hidden states can distort fine-grained cues such as texture, color, and spatial layout. Effective visual latent reasoning requires a visual manifold that can preserve rich perceptual evidence while remaining compatible with language-grounded reasoning.

\noindent \textbf{Latent Visual Reasoning.} 
To move beyond static perception toward internal visual imagination, recent paradigms have explored performing logical deductions directly within the latent space. LVR\citep{li2025latent} performs autoregressive reasoning within the visual embedding space by reconstructing task-critical tokens from latent states. Monet generates continuous embeddings that serve as intermediate visual thoughts and aligns them with the visual semantic space through a distillation pipeline\citep{wang2025monet}. Mirage further treats hidden states as latent visual tokens to build multimodal reasoning trajectories without pixel-level image synthesis\citep{yang2025machine}. Despite these advances, recent diagnostic studies reveal a persistent causality gap: \textit{latent tokens are often weakly grounded in visual inputs and exert limited influence on final answers} \citep{li2026imagination}. Our analyses further reveal a fundamental \textit{manifold divergence} in existing baselines, where latent trajectories drift into deep, uncalibrated regions far from the pre-trained semantic anchors. These limitations thus motivate our grounded latent reasoning framework.
\vspace{-6mm}
\section{Analysis on Reasoning Manifold}
\vspace{-2mm}
To understand how latent tokens shape the reasoning trajectory of models trained for latent reasoning, we develop a geometric analysis that visualizes the path traversed by hidden states during a single inference, relative to both the original
base-model manifold and the vocabulary embedding space. The analysis is motivated by a simple question: \textit{as the model generates a sequence of latent tokens followed by the decoded language answer tokens, how does its internal representation travel through the joint space of hidden states and vocabulary embeddings?}


We construct a dataset of reasoning trajectories from an evaluation set of $N$ samples. For each sample $i$, a forward-decoding pass produces last-layer hidden states $\{\mathbf{h}^{(i)}_t\}_{t=1}^{T_i}$, where each state is labeled as belonging to either the \emph{latent} or \emph{answer} phase. We denote the aggregate of these reasoning states across all samples as $\mathcal{H}_{\textit{RIS}}$. As references, we collect the corresponding hidden states from a frozen base model, denoted as $\mathcal{H}_{\mathrm{base}}$, alongside the vocabulary embedding matrix $\mathbf{E} \in \mathbb{R}^{V \times d}$. To visualize manifold distributions and reasoning trajectories in a shared space, we fit PCA jointly on $\mathcal{H}_{\mathrm{base}}$, $\mathcal{H}_{\textit{RIS}}$, and $\mathbf{E}$, and project each trajectory onto the plane spanned by the leading two principal components.

\begin{figure}
    \centering
    \includegraphics[width=1\linewidth]{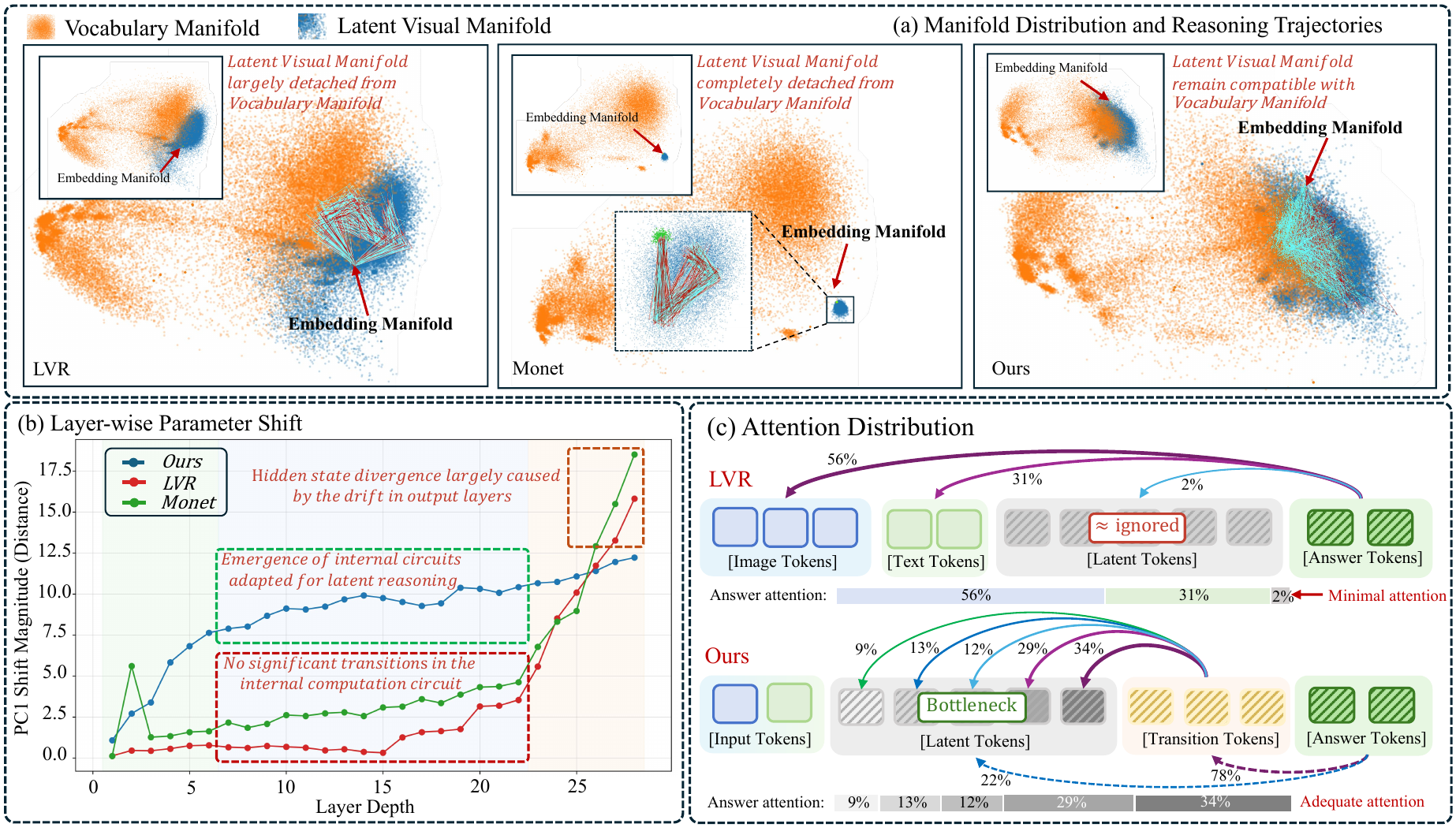}
    \caption{Geometric analysis of latent reasoning paradigms: (a) manifold distribution and trajectories, (b) layer-wise parameter shift relative to the base model, and (c) attention pattern of answer tokens.}
    \label{fig:manifold-traj}
\end{figure}

\subsection{Manifold Compatibility and Trajectory Dynamics}
We use \emph{manifold} to refer to the empirical support of high-dimensional hidden-state or embedding representations. Figure \ref{fig:manifold-traj} compares the hidden-state distributions of the frozen base model as an empirical reference for the pretrained vocabulary-aligned manifold with those induced by different latent reasoning training methods. In LVR and Monet, the learned latent states are visibly separated from this reference manifold, suggesting that they form distinct latent visual manifolds with richer visual expressiveness but also introduce representational distribution shifts. Such separation can weaken compatibility with the pretrained reasoning circuits acquired during large-scale language pretraining and with the language decoding process, which partly explains their degraded performance. 

The trajectory visualization provides a dynamic view of this phenomenon. Successful reasoning paths tend to remain connected to the vocabulary-aligned manifold, whereas failed paths are more often trapped within detached latent visual regions. This does not imply that correct reasoning must explicitly return to the vocabulary manifold at specific steps; rather, effective latent visual reasoning should remain compatible with the pretrained representation regime, allowing the model to exploit existing reasoning circuits while incorporating fine-grained visual evidence in latent space. This supports our view: \textit{latent visual reasoning should not replace the model’s original reasoning manifold, but should develop as a compatible extension of it.}

\begin{figure}
    \centering
    \includegraphics[width=1\linewidth]{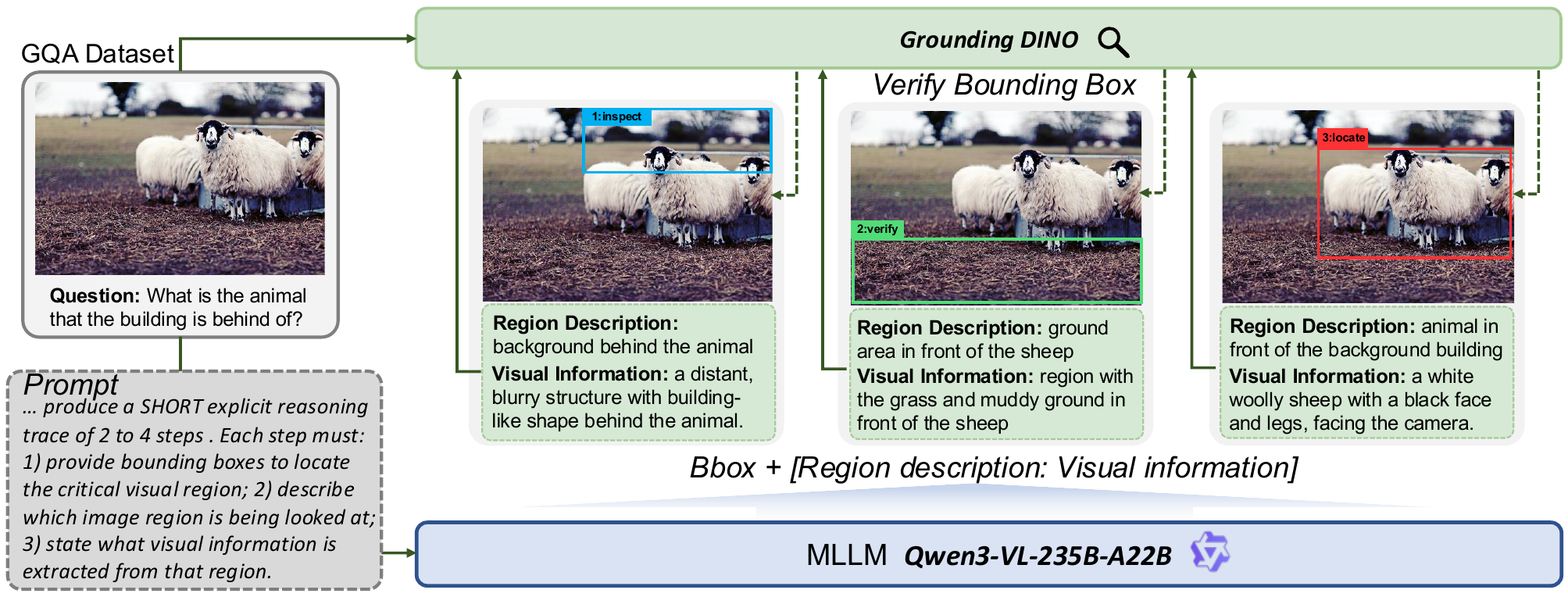}
    \caption{Dataset construction pipeline. An MLLM decomposes each QA pair into several grounded reasoning steps, which are then verified and calibrated by Grounding DINO.}
    \label{fig:placeholder}
\end{figure}

\subsection{Layer-wise Adaptation Pattern}
To further analyze the observed manifold compatibility, Figure \ref{fig:manifold-traj}(b) measures the layer-wise parameter shift from the base model. LVR and Monet show limited changes in the middle layers but large shifts in the output layers, indicating that their adaptation is concentrated near the final decoding interface rather than distributed across the internal computation stack. This pattern suggests that they form limited internal circuits for latent visual reasoning and instead rely on late-stage compensation to map detached latent visual states back to language-grounded outputs. In contrast, our method produces a smoother and more sustained shifts across the middle layers, followed by much milder changes near the output layers. This indicates that the model gradually adapts its internal computation to support latent visual reasoning while preserving the vocabulary-aligned manifold near the decoding interface. 

\subsection{Latent Bypassing and Trajectory Collapse}
The attention analysis further explains why a detached latent visual manifold does not necessarily lead to effective latent reasoning. As shown in Figure \ref{fig:manifold-traj}(c), answer tokens allocate most of their attention to original image and text tokens, while assigning only minimal attention to latent tokens. This indicates that the final decoding largely bypasses the latent reasoning tokens instead of using them as intermediate computational states. This observation is consistent with \citep{li2026imagination} that latent tokens are weakly grounded in the visual premises and exert limited causal influence on the final answer. The trajectories of LVR and Monet in Figure \ref{fig:manifold-traj}(a) provide a complementary view. Across samples, their latent trajectories are highly similar and densely collapsed, suggesting limited instance-specific reasoning information. Thus, although these methods appear to form latent visual manifolds, such manifolds are not effectively integrated into the model’s computation: they neither receive sufficient input-grounded variation nor provide a reliable pathway back to the vocabulary-aligned manifold. 

Taken together, these analyses suggest: \textit{Effective latent reasoning requires more than learning a separated latent visual manifold. It must establish compatible internal circuits that ground latent states in visual inputs, preserve access to pretrained reasoning circuits, and smoothly transfer visual abstractions back to the vocabulary-aligned manifold for language-grounded answer generation.}



\begin{figure}
    \centering
    \includegraphics[width=1\linewidth]{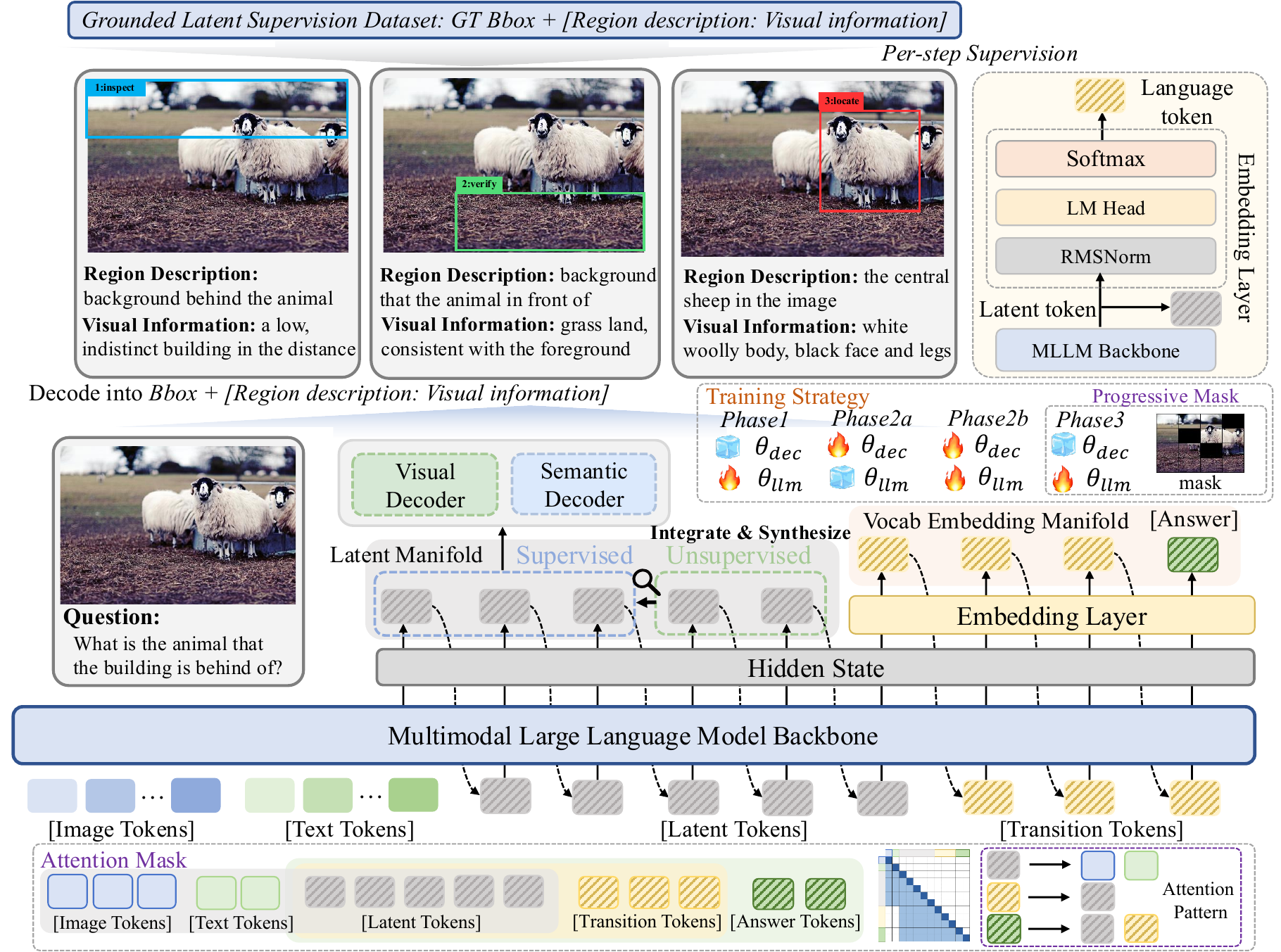}
    \caption{Overview of \textbf{\textit{RIS}} framework. Visual and semantic decoders are only used for supervision in training and will be removed during inference. Full attention mask used during inference.}
    \label{fig:main-fig}
\end{figure}

\section{Method}
\label{sec:method}



In this section, we first elaborate on the construction of our step-wise \textit{Grounded Latent Supervision Dataset (GLSD)}, which provides spatial-semantic supervision for intermediate visual reasoning. Built on this, we introduce \textit{\textbf{RIS}}, a grounded latent reasoning framework that uses last-layer hidden states as continuous latent tokens. Following common practice, \textbf{\textit{RIS}} allocates a fixed number of latent tokens and feeds them forward as continuous embeddings. Figure~\ref{fig:main-fig} illustrates the overall training pipeline.

\subsection{Dataset Construction}
\label{sec:dataset}

Since existing datasets rarely provide fine-grained supervision for intermediate reasoning, we convert standard GQA\citep{Hudson2019GQAAN} data into a unified step-wise grounded format and curate such supervision through a two-stage pipeline. First, prompt an MLLM to decompose each question-answer pair into a 2--4 step trace, where each step contains an operation tag, a region-specific semantic description, visual information, and preliminary bounding-box coordinates, leading to a full answer (as transition tokens) and final answer as visualized in Appendix~\ref{dataset_viz}. Then, we feed the generated descriptions and original image into Grounding DINO~\citep{liu2023grounding} to verify and calibrate the MLLM-proposed coordinates.

The verified trace is serialized as $\mathcal{T}$ and padded to the fixed latent budget $K$. Each supervised step is represented as \texttt{<step\_start> <bbox> $b_i$ </bbox> $v_i$ <step\_end>}, where $b_i$ denotes the normalized bounding box coordinates and $v_i$ denotes the corresponding visual information. Let $m_i\in\{0,1\}$ indicate whether the $i$-th slot has explicit step-wise supervision. For unsupervised slots with $m_i=0$, we insert \texttt{<step\_start> [synthesize] <step\_end>}, allowing them to learn evidence integration from downstream answer supervision. This yields $96k$ training samples.

\subsection{Training Phase 1: Explicit Grounded Reasoning Initialization}

In Phase 1, we initialize the MLLM with explicit grounded visual reasoning before moving to continuous latent computation. Given image embeddings $\mathcal{X}_v$ and a tokenized query $\mathcal{X}_q$, the model is trained to generate the structured reasoning trace $\mathcal{T}$ followed by the \textit{full answer} $A_s$ and final answer $A$. Unlike generic textual rationales, $\mathcal{T}$ contains step-wise normalized bounding-box coordinates, fine-grained visual information, and boundary tokens, thereby teaching the model to retrieve regions, integrate visual evidence, and synthesize the answer in a discrete but explicitly grounded format.

We optimize the backbone $\mathcal{M}$ with the standard next-token prediction (NTP) objective:

\begin{equation}
\small
\mathcal{L}_{\mathrm{NTP}}
=
-\sum_{t=1}^{|\mathcal{T}\oplus A_s \oplus A|}
\log P_{\mathcal{M}}\left(y_t \mid y_{<t}, \mathcal{X}_v, \mathcal{X}_q\right).
\end{equation}

This stage equips the backbone with a reliable text-space grounded reasoning prior. After training, we use the Phase-1 trained model to encode the region-specific descriptions in each step. The resulting cached embeddings serve as stable semantic anchors for Phase 2 training, preventing the latent-space supervision targets from drifting during continuous-token training. 

\subsection{Training Phase 2a: Side-Head Grounding and Semantic Alignment}

To transition from discrete textual reasoning to the continuous latent space, we introduce a set of fixed-capacity visual latent tokens $\mathcal{C} = \{c_1, c_2, \dots, c_K\}$. In this phase, we freeze the MLLM backbone and calibrate two lightweight, task-specific side heads: \textit{Visual Grounding Decoder} ($f_{\text{reg}}$) and \textit{Semantic Alignment Decoder} ($f_{\text{desc}}$). For each reasoning step $t$ where $m_t=1$, we extract the last-layer hidden state $z_t$ at the \texttt{<step\_start>} token in the explicit reasoning trace, which summarizes the context before generating the step content. The two decoders then operate as explicit supervision signals:

\noindent \textbf{Spatial Grounding:} $f_{\text{reg}}$ predicts the bounding box coordinates $\hat{b}_t = f_{\text{reg}}(z_t)$. We optimize this using a combination of $\ell_1$ distance and Generalized IoU (GIoU) loss against the verified boxes $b_t$:
\begin{equation}
\small
\mathcal{L}_{\text{reg}} = \frac{1}{\sum_{t=1}^K m_t} \sum_{t=1}^K m_t \Big( ||\hat{b}_t - b_t||_1 + \lambda_{\text{GIoU}} \mathcal{L}_{\text{GIoU}}(\hat{b}_t, b_t) \Big)
\end{equation}
    
\noindent \textbf{Semantic Anchoring:} To prevent the latent representation from drifting into uninterpretable regions, $f_{\text{desc}}$ projects $z_t$ into a semantic embedding space. We minimize the cosine distance between the projected vector and the pre-computed text embeddings $e_t$ of the corresponding region description:

\begin{equation}
\small
\mathcal{L}_{\text{desc}} = \frac{1}{\sum_{t=1}^K m_t} \sum_{t=1}^K m_t \Big( 1 - \cos(f_{\text{desc}}(z_t), e_t) \Big)
\end{equation}

The total loss for this warm-up phase is $\mathcal{L}_{\text{Phase2a}} = \lambda_r \mathcal{L}_{\text{reg}} + \lambda_d \mathcal{L}_{\text{desc}}$. This supervised calibration imposes spatial grounding to specify \emph{where to look}, while semantic anchoring to specify \emph{what is seen}. These calibrated side heads provide stable spatial-semantic constraints for converting explicit reasoning steps into grounded latent tokens in the subsequent phase.

\subsection{Training Phase 2b: Progressive Latent Internalization}


With the side heads calibrated, Phase 2b progressively converts explicit textual reasoning into continuous latent computation. Following the \textit{Coconut} curriculum paradigm~\citep{hao2024training}, we use a step-wise replacement schedule $s\in\{1,\dots,K\}$: at stage $s$, the first $s$ textual reasoning blocks are replaced by their corresponding latent states $\mathcal{C}_{\le s}$, while the remaining textual steps $\mathcal{T}_{>s}$ and the \textit{full answer} $A_s$ and final answer $A$ are still generated autoregressively as discrete text. Unlike Phase 2a, the MLLM backbone is unfrozen, allowing its internal computation to adapt to latent inputs.


The model is trained with next-token prediction on the remaining textual trace and final answer:

\begin{equation}
\small
\mathcal{L}_{\text{cot}} = - \sum_{t=1}^{|\mathcal{T}_{>s}|} \log P_{\mathcal{M}}(y_t \mid y_{<t}, \mathcal{X}_v, \mathcal{X}_q, \mathcal{C}_{\le s})
, \hspace{2mm}
\mathcal{L}_{\text{ans}} = - \sum_{t=1}^{|A_s \oplus A|} \log P_{\mathcal{M}}(a_t \mid a_{<t}, \mathcal{X}_v, \mathcal{X}_q, \mathcal{C}_{\le s}, \mathcal{T}_{>s})
\end{equation}

To preserve grounding during internalization, the side-head losses are applied only to supervised latent slots that have been replaced, i.e., $\{t\le s \mid m_t=1\}$. The overall objective is

\begin{equation}
\mathcal{L}_{\text{Phase2b}} = \mathcal{L}_{\text{ans}} + \alpha \mathcal{L}_{\text{cot}} + \lambda_r \mathcal{L}_{\text{reg}} + \lambda_d \mathcal{L}_{\text{desc}}
\end{equation}


As $s$ increases, information previously expressed by text is gradually internalized into continuous latent states. By the end of this curriculum, the model learns to retrieve and transform grounded visual evidence in latent space with reduced dependence on explicit verbalization.

\subsection{Training Phase 3: Bottlenecked Latent Integration}



In Phase 3, all explicit reasoning steps are replaced by latent tokens, so the intermediate reasoning process is fully mediated by the latent tokens $\mathcal{C}$. To prevent answer decoding from bypassing these tokens and directly attending to the original image and query tokens $(\mathcal{X}_v,\mathcal{X}_q)$, we introduce a \textit{Progressive Attention Mask}. Specifically, we anneal a masking probability $\rho(\tau)$ over training step $\tau$ and sample $M\sim\mathrm{Bernoulli}(\rho(\tau))$ to modulate the causal attention matrix. As $\rho(\tau)$ increases, answer tokens are gradually forced to rely on $\mathcal{C}$, making latent states the information conduit for task-relevant visual and semantic evidence.

Since no textual reasoning steps remain in this phase, the textual CoT loss is removed. The model is trained with the final-answer objective, together with side-head constraints on supervised latent slots:

\begin{equation}
\mathcal{L}_{\text{Phase3}} = \mathcal{L}_{\text{ans}} + \lambda_r \mathcal{L}_{\text{reg}} + \lambda_d \mathcal{L}_{\text{desc}}
\end{equation}

Under the fully masked condition, the answer loss is conditioned on the latent tokens:

\begin{equation}
\small
\mathcal{L}_{\mathrm{ans}}
=
-\sum_{t=1}^{|A_s\oplus A|}
\log P_{\mathcal{M}}\left(u_t \mid u_{<t}, \mathcal{C}\right),
\end{equation}

where $A_s$ and $A$ denote the \textit{full answer} bridge and the final answer,  $u_t$ indexes concated tokens.

Under this information bottleneck, the supervised latent tokens remain spatially and semantically grounded through the side-head losses, while the free tokens, which receive no direct spatial-semantic supervision, are optimized only through the answer objective and are therefore encouraged to aggregate and synthesize the evidence retrieved by earlier grounded tokens.

Finally, to facilitate a smooth return from the latent visual manifold to the discrete vocabulary manifold $\mathcal{M}_{vocab}$, we utilize the short \textit{full answer} $A_s$  (as defined in Phase 1) as a sequence of \textit{Manifold Transition Tokens}. Instead of forcing an abrupt jump from synthesized latent states to final answer decoding, $A_s$ provides an autoregressive intermediate path that gradually maps latent visual representations back toward the pretrained vocabulary-aligned manifold $\mathcal{M}_{vocab}$. Its dense next-token supervision stabilizes this transition and improves the compatibility between latent visual reasoning and language-grounded answer generation.

\section{Experiments}
\vspace{-2mm}
\subsection{Experiment Setup}
\label{experiment_setup}
\vspace{-1mm}

\textbf{Training and Evaluation Setup.} 
We adopt Qwen2.5-VL-7B\citep{bai2025qwen25vltechnicalreport} as our base model. The training process follows our proposed three-phase pipeline and all parameters are detailed in Appendix~\ref{hyperparameters}.

\textbf{Evaluated Benchmarks.} 
To comprehensively evaluate our proposed method, we conduct experiments on a diverse set of challenging perception and reasoning benchmarks: $\text{V}^*$\citep{10657571}, HRBench4K\citep{wang2025divide}, HRBench8K\citep{wang2025divide}, MMVP\citep{tong2024eyes}, and BLINK\citep{fu2024blink}.

\textbf{Baselines.} 
We compare \textbf{\textit{RIS}} against a variety of baselines: (1) \textit{Proprietary Models:} GPT-4o; (2) \textit{Open-Source Base Model:} Qwen2.5-VL-7B; (3) \textit{Vanilla SFT:} Qwen2.5-VL-7B+GLSD, which finetunes the base model on our curated \textit{Grounded Latent Supervision Dataset (GLSD)} for same training steps as \textbf{\textit{RIS}}; (4) \textit{Latent Visual Reasoning Methods:} LVR\citep{li2025latent}, Monet\citep{wang2025monet}, and CoVT\citep{qin2025chain}. Furthermore, to investigate the benefit of reinforcement learning on our method, we introduce \textit{\textbf{\textit{RIS}}+VLPO}, a variant that further optimizes our model using \textit{Visual-latent Policy Optimization (VLPO)} proposed by Monet.

\subsection{Main Results} 

Table~\ref{tab:main-table} summarizes the performance of \textbf{\textit{RIS}} and baselines across five visual reasoning benchmarks. Overall, \textbf{\textit{RIS}} consistently outperforms both the open-source backbone and existing latent visual reasoning baselines. The GLSD baseline, which retains explicit textual reasoning, yields much smaller gains. This indicates that the improvements are not only due to extra supervision but mainly stem from internalizing grounded visual evidence into latent states and performing latent reasoning. Although \textbf{\textit{RIS+VLPO}} does not consistently yield further gains, it remains a promising direction for adaptive and stable latent computation, while a reliable latent variant is still underexplored.

\begin{wraptable}{r}{0.48\textwidth}
\centering
\resizebox{0.48\textwidth}{!}{ 
\begin{tabular}{lccccc}
\toprule
\textbf{Model} & \textbf{S.R.} & \textbf{O.L.} & \textbf{R.R.} & \textbf{Counting} & \textbf{R.D.} \\
\midrule
Base & 86.81 & 43.62 & 39.87 & 68.30 & 67.26 \\
Base+\textbf{\textit{GLSD}} & 87.03 & 49.57 & 41.13 & 67.59 & 65.66 \\
\midrule
LVR & 86.01 & 50.82 & 43.28 & 69.17 & 76.61 \\
Monet & 85.31 & 45.08 & 39.55 & 70.83 & 75.81 \\
CVOT & 87.41 & 53.20 & 38.06 & 65.83 & \textbf{77.65} \\
\midrule
\textbf{\textit{RIS}} & \textbf{89.60} & \textbf{54.63} & \textbf{44.25} & \textbf{72.02} & 76.50 \\
\textit{Improvement} & \textcolor{green!60!black}{+2.79} & \textcolor{green!60!black}{+11.01} & \textcolor{green!60!black}{+4.38} & \textcolor{green!60!black}{+3.72} & \textcolor{green!60!black}{+9.24} \\
\bottomrule
\end{tabular}
}
\caption{Performance on \textit{BLINK}. \textit{(S.R.: Spatial Reasoning, O.L.: Object Localization, R.R.: Relative Reflectance, R.D.: Relative Depth).}}
\label{tab:blink-table}
\vspace{-3mm}
\end{wraptable}

\begin{table*}[t]
\centering
\resizebox{\textwidth}{!}{
\begin{tabular}{lccccccccccc}
\toprule
\multirow{2}{*}{\textbf{Model}} 
& \multicolumn{3}{c}{\textbf{V*}} 
& \multicolumn{3}{c}{\textbf{HRBench4K}} 
& \multicolumn{3}{c}{\textbf{HRBench8K}} 
& \multirow{2}{*}{\textbf{MMVP}}
& \multirow{2}{*}{\textbf{BLINK}} \\
\cmidrule(lr){2-4}
\cmidrule(lr){5-7}
\cmidrule(lr){8-10}
& Overall & Attribute & Spatial
& Overall & FSP & FCP
& Overall & FSP & FCP
& & \\
\midrule

\multicolumn{12}{c}{\cellcolor{orange!10}\textbf{\textit{Proprietary Model}}} \\
GPT-4o
& 65.15 & 69.68 & 58.39
& 54.70 & 64.93 & 44.51
& 51.12 & 57.08 & 45.12
& 72.00 & 63.55 \\

\midrule
\multicolumn{12}{c}{\cellcolor{blue!10}\textbf{\textit{Open-Source Model}}} \\
Qwen2.5-VL-7B
& 76.65 & 77.12 & 74.35
& 68.30 & 80.60 & 56.03
& 64.33 & 74.42 & 54.24
& 63.49 & 54.94 \\

Qwen2.5-VL-7B+\textbf{\textit{GLSD}}
& 78.25 & 78.39 & 78.11
& 70.58 & 83.30 & 57.86
& 64.70 & 74.69 & 54.70
& 66.80 & 56.25 \\

LVR\citep{li2025latent}
& 80.60 & 83.26 & 77.94
& 70.88 & 83.25 & 57.50
& 63.50 & 75.00 & 52.00
& 71.03 & 56.79 \\

Monet\citep{wang2025monet}
& 81.90 & 81.94 & 81.86
& 69.97 & 84.01 & 55.93
& 67.01 & 78.59 & 55.43
& 70.00 & 56.70 \\

COVT\citep{qin2025chain}
& 79.10 & 81.05 & 77.15
& 71.90 & 85.50 & 58.30
& 68.40 & \textbf{79.30} & 57.50
& 58.70 & 57.40 \\

\midrule
\multicolumn{12}{c}{\cellcolor{green!8}\textbf{\textit{Our Model}}} \\
\textbf{\textit{RIS}}
& \textbf{83.75} & \textbf{84.26} & \textbf{83.24}
& \textbf{73.23} & \textbf{86.33} & 60.12
& \textbf{68.52} & 79.05 &  \textbf{57.98}
& 73.55 & 60.60 \\

\textbf{\textit{RIS}+\textit{VLPO}\citep{wang2025monet}}
& 81.76 & 81.24 & 82.28
& 71.79 & 82.83 & \textbf{60.75}
& 63.05 & 72.67 & 53.42
& \textbf{73.76} & \textbf{60.95} \\

\textit{Relative Improvement}
& \textcolor{green!60!black}{+7.10}
& \textcolor{green!60!black}{+7.14}
& \textcolor{green!60!black}{+8.89}
& \textcolor{green!60!black}{+4.93}
& \textcolor{green!60!black}{+5.73}
& \textcolor{green!60!black}{+4.09}
& \textcolor{green!60!black}{+4.19}
& \textcolor{green!60!black}{+4.63}
& \textcolor{green!60!black}{+3.74}
& \textcolor{green!60!black}{+10.27}
& \textcolor{green!60!black}{+5.66} \\

\bottomrule
\end{tabular}
}
\caption{Main results on visual reasoning benchmarks across proprietary, open-source, and latent visual reasoning baselines (with Qwen2.5-VL-7B as the same backbone) with \textbf{5 latent tokens}.}
\label{tab:main-table}
\vspace{-6mm}
\end{table*}

The performance pattern further aligns with the characteristics of each benchmark. \textbf{\textit{RIS}} achieves larger gains on benchmarks requiring precise localization, structured visual search, and multi-step perceptual reasoning, such as V$^\ast$ and BLINK. The detailed BLINK results in Table~\ref{tab:blink-table} show particularly strong improvements on tasks closely tied to visual grounding and sequential evidence retrieval, including \textit{Spatial Reasoning}, \textit{Object Localization}, and \textit{Relational Reasoning}. In contrast, on MMVP, where performance is more constrained by the visual encoder's ability, the gains are more moderate.

\subsection{Analysis on Latent Behaviors}

\paragraph{Impact of Latent Token Budget.} Figure~\ref{fig:ablation-perf}(c) studies the effect of latent token budget. Although more latent tokens provide additional computation, performance does not improve monotonically. This is because most training samples contain only three to four supervised reasoning steps, which already match the typical length of grounded logical reasoning. Tokens beyond this range receive no spatial-semantic supervision and only learn evidence integration from the final-answer objective.

The effect is therefore task-dependent. For benchmarks requiring multi-step visual exploration, such as V$^\ast$ and BLINK, a small number of extra tokens can help integrate retrieved evidence. However, larger budgets introduce more unsupervised slots, weakening grounding reliability and increasing optimization instability. This degradation is more evident on HRBench8K and MMVP: the former requires highly reliable local evidence under extreme resolution, while the latter is mainly limited by the visual encoder's ability. Overall, the latent budget should match the steps of available supervision, enabling \textbf{\textit{RIS}} to expand latent reasoning without diluting grounded visual evidence.

\vspace{-4mm}

\begin{figure}
    \centering
    \includegraphics[width=1\linewidth]{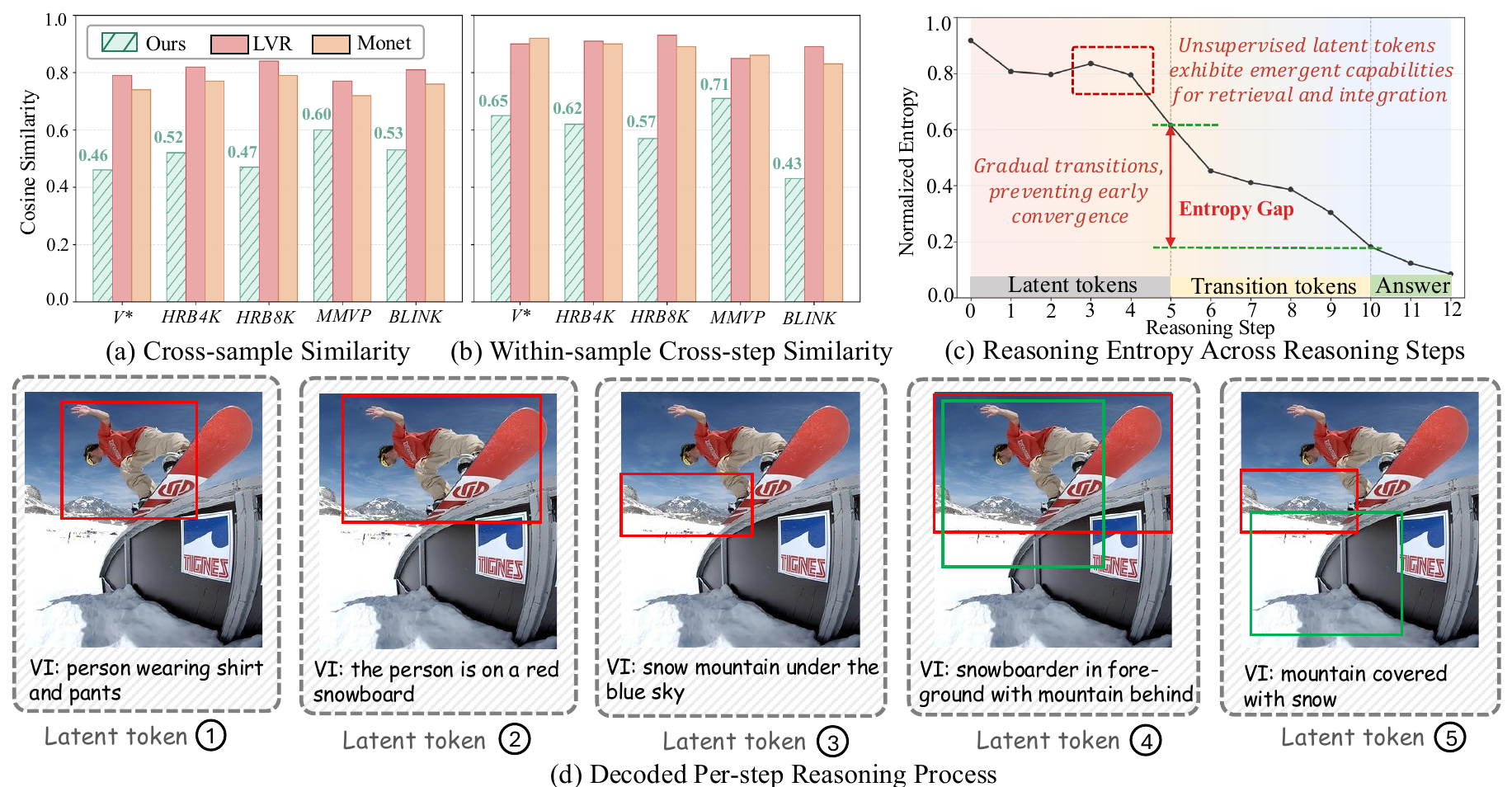}
    \vspace{-4mm}
    \caption{Latent Behavior Analysis of \textbf{\textit{RIS}}: Diversity, Reasoning Entropy, and Interpretability. }
    \label{fig:exp-main-fig}
    \vspace{-4mm}
\end{figure}

\paragraph{Entropy Dynamics of Latent Reasoning.}
Figure~\ref{fig:exp-main-fig}(c) visualizes the normalized reasoning entropy along the latent trajectory, with details provided in the Appendix~\ref{app:reasoning-entropy}. The supervised latent tokens maintain high entropy, suggesting they preserve an open visual reasoning space rather than being constrained to a single reasoning path. The unsupervised slots show a slight entropy increase, indicating their role in aggregating and synthesizing evidence retrieved by earlier grounded tokens. Following latent tokens, transition tokens bridge high-entropy latent states to low-entropy answer tokens, avoiding an abrupt representation jump. This supports our design: \textit{visual evidence is first explored and integrated in the latent space, then transited back to the vocabulary decoding space.}

\paragraph{Latent Token Diversity and Interpretability.}
Figure~\ref{fig:exp-main-fig}(a)(b) examines whether latent tokens collapse into instance-agnostic patterns. In both cross-sample and within-sample analyses, \textbf{\textit{RIS}} shows much lower similarity than LVR and Monet, indicating more step-specific and instance-dependent latent states. Although within-sample similarity is naturally higher due to shared visual content within the same image, \textbf{\textit{RIS}} remains notably diverse, suggesting that its latent tokens progressively organize grounded visual-semantic evidence rather than repeatedly encoding the same features. This effect is especially clear on BLINK, where repeated visual search and structured perception are required. Cross-sample similarity is overall lower, and the remaining similarity is likely due to similarity across examples; nevertheless, \textbf{\textit{RIS}} still maintains substantially stronger diversity than prior methods.

Figure~\ref{fig:exp-main-fig}(d) further provides a qualitative view. The decoded bounding boxes form a clear step-by-step reasoning trajectory, and the associated visual information captures key evidence in each region. The final synthesis tokens cover multiple semantically relevant regions and decode integrated visual information rather than isolated local details. This confirms their role as an evidence integration stage, consistent with the entropy increase observed in Figure~\ref{fig:exp-main-fig}(c).

\subsection{Ablation Study}

\begin{figure}
    \centering
    \includegraphics[width=1\linewidth]{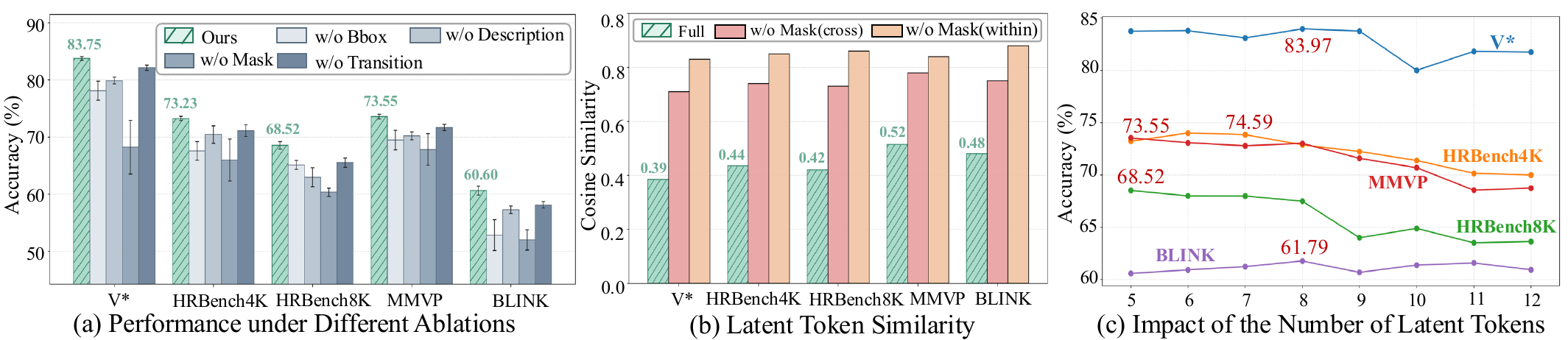}
    \caption{Design Ablations. {\small (Error bars denote accuracy standard deviation across repeated training runs.)}}
    \label{fig:ablation-perf}
\end{figure}


To assess the contribution of each component in \textbf{\textit{RIS}}, we conduct systematic ablations, with results summarized in Figure~\ref{fig:ablation-perf} leading to three critical takeaways regarding latent visual reasoning.

\textbf{Spatial-semantic supervision stabilizes grounded latent reasoning.}
Removing either side-head supervision degrades performance and increases variance, confirming the importance of explicit grounding. Bounding-box supervision is especially important on localization-heavy benchmarks such as V$^\ast$ and BLINK, showing that spatial anchoring is crucial for directing latent tokens to task-relevant visual evidence. Description supervision has a milder but consistent impact, suggesting that semantic alignment mainly regularizes the latent space and prevents drift from the vocabulary reasoning manifold.

\textbf{The progressive attention mask prevents latent bypassing and collapse.}
Removing the progressive attention mask causes the largest performance drop. Without this bottleneck, answer tokens can directly attend to the raw image and question, weakening the role of latent tokens. The degradation further suggests that uncalibrated latent tokens are not merely ignored placeholders; they can interfere with prediction even when answer tokens still access the original inputs. This is further supported by Figure~\ref{fig:ablation-perf}(b), removing the mask also sharply increases both cross-sample and within-sample latent-token similarity, revealing collapse toward generic representations. Thus, the mask is essential for routing visual evidence through the latent trajectory and maintaining step-specific latent computation.

\textbf{Transition tokens facilitate language-grounded decoding.}
Removing transition tokens also consistently hurts performance, indicating that the short-answer bridge is important for mapping synthesized latent evidence back to language-grounded decoding. This supports our hypothesis that transition tokens reduce representation mismatch between the expressive latent visual manifold and the vocabulary-aligned output space.

\section{Conclusion}

In this work, we propose \textbf{\textit{RIS}}, a spatial-semantic grounded latent visual reasoning framework that enables MLLMs to reason over visual evidence within continuous latent states. We show that existing latent visual reasoning methods suffer from weak grounding, trajectory collapse, and answer shortcuts. To address these issues, \textbf{\textit{RIS}} anchors latent tokens with spatial-semantic supervision, enforces their use through a progressive attention bottleneck, and introduces transition tokens to bridge latent reasoning back to language decoding phase. Extensive experiments  demonstrate consistent improvements over strong baselines. Further analyses verify that \textbf{\textit{RIS}} learns more diverse, interpretable, and causally effective latent trajectories, suggesting a practical path to faithful internal visual reasoning in MLLMs.

\bibliographystyle{unsrtnat}
\bibliography{references}

\newpage

\appendix

\section{Limitations}
\label{limitations}

Although \textbf{\textit{RIS}} provides an effective framework for grounded latent visual reasoning, it still has several limitations.

First, our framework involves multiple interacting components. 
While each component is empirically essential, the overall framework introduces several hyperparameters and schedules whose optimal settings may depend on the backbone model, task type, and data distribution.  A more principled and automatic strategy for configuring these components would further improve the robustness and usability of our framework.

Second, \textbf{\textit{RIS}} still fixes the latent token budget, which is a common setting of most existing latent reasoning methods. In our framework, the supervised latent tokens and unsupervised latent tokens provide an implicit form of adaptive computation, and our analyses suggest that different latent slots can spontaneously specialize into retrieval, integration, and synthesis roles. Nevertheless, the total number of latent slots is still predefined rather than dynamically determined according to the complexity of each input. Developing latent reasoning models with truly adaptive token allocation remains an important direction of our future work.

Third, our experiments mainly focus on standard image-based VQA and visual reasoning benchmarks commonly used for latent visual reasoning. Although these benchmarks cover localization, structured visual search, and multi-step perceptual reasoning, they do not fully reflect the diversity of real-world multimodal reasoning scenarios. Future work would further evaluate and extend \textbf{\textit{RIS}} to broader settings, such as open-ended long-form visual question answering, multi-image reasoning, video reasoning, and interactive embodied tasks.

\section{Training Hyperparameters}
\label{hyperparameters}

\begin{table}[h] \small
\centering
\caption{Detailed Hyperparameters for \textbf{\textit{RIS}} Training.}
\label{tab:hyperparameters}
\begin{tabular}{llc}
\toprule
\textbf{Training Phase} & \textbf{Hyperparameter} & \textbf{Value} \\
\midrule
\multirow{2}{*}{Architecture} 
& Base Model & Qwen2.5-VL-7B \\
& Latent Tokens ($K$) & 5 \\
\midrule
\multirow{2}{*}{Phase 1} 
& Epochs & 1 -- 2 \\
& Learning Rate & $1 \times 10^{-5}$ \\
\midrule
\multirow{4}{*}{Phase 2} 
& Sub-stage 2.A LR ($f_{reg}$) & $5 \times 10^{-4}$ \\
& Sub-stage 2.A LR ($f_{desc}$) & $1 \times 10^{-4}$ \\
& Sub-stage 2.B Epochs per Curriculum Stage & 1 -- 2 \\
& Sub-stage 2.B LR & $5 \times 10^{-6}$ \\
\midrule
\multirow{3}{*}{Phase 3} 
& Epochs & 1 -- 2 \\
& Learning Rate & $3 \times 10^{-6}$ \\
& Mask Ratio Annealing & $0.3 \rightarrow 1.0$ \\
\midrule
\multirow{4}{*}{Loss Weights} 
& $\lambda_r$ (Region Grounding) & 1.0 \\
& $\lambda_{giou}$ (GIoU within Region Grounding) & 2.0 \\
& $\lambda_d$ (Region Description Alignment) & 0.5 \\
& $\alpha$ (Textual CoT) & 1.0 \\
\bottomrule
\end{tabular}
\end{table}

\begin{table}[t]
\centering
\small
\caption{Additional hyperparameters for the \textbf{\textit{RIS}}+VLPO reinforcement learning stage.}
\label{tab:ris-vlpo-hparams}
\begin{tabular}{ll}
\toprule
Hyperparameter & Value \\
\midrule
Initialization & Phase-3 \textbf{\textit{RIS}} checkpoint \\
RL data & 3.2k/\textbf{6.4k} samples from GLSD \\
Latent action & \textbf{\textit{RIS}} latent slot hidden state $c_t$ \\
Latent tokens & $K=5$ \\
Attention mask & Phase-3 bottleneck mask, fixed at $\rho=1.0$ \\
Reference model & Frozen Phase-3 \textbf{\textit{RIS}} checkpoint \\
Rollouts per prompt & $G=4$ \\
Policy epochs per batch & 1 \\
RL epochs & 1 \\
Global prompt batch size & 32 \\
Learning rate & $1\times10^{-6}$ \\
KL coefficient & $\beta=0.02$ \\
Latent likelihood scale & $\sigma=1.0$ \\
Max response length & 128 tokens \\
Sampling temperature / top-$p$ & 1.0 / 0.95 \\
Reward & $r_{\rm acc}+0.1r_{\rm fmt}$ \\
Accuracy reward $r_{\rm acc}$ & 1 if normalized answer matches ground truth, else 0 \\
Format reward $r_{\rm fmt}$ & 1 if final answer is extractable, else 0 \\
Latent-use reward & None \\
Grounding regularization & $0.1\mathcal{L}_{\rm reg}+0.05\mathcal{L}_{\rm desc}$ \\
Trainable modules & MLLM backbone; side heads frozen \\
\bottomrule
\end{tabular}
\end{table}

\section{Compute Resources}
\label{compute_resources}

All experiments were conducted on a single server equipped with an Intel Xeon Platinum 8383C CPU, 
512GB DDR5 RAM, and 4 NVIDIA A100 80GB GPUs connected with NVLink. 
We trained on an 80k-sample \textit{Grounded Latent Supervision Dataset} (GLSD) with a global batch size of 32, corresponding to 2,500 optimization steps for each full data pass. 
Table~\ref{tab:compute_resource} reports the estimated wall-clock training time for each stage of \textbf{\textit{RIS}}. 
For Phase 2b, we assume a latent budget of $K=5$ and one full data pass for each curriculum stage; its cost scales approximately linearly with $K$.

\begin{table}[t]
\centering
\small
\caption{Recorded training time of different stages on 4 NVIDIA A100 80GB GPUs.}
\label{tab:compute_resource}
\begin{tabular}{lccc}
\toprule
\textbf{Training Stage} & \textbf{Trainable Components} & \textbf{Update Steps} & \textbf{Wall-clock Time} \\
\midrule
Phase 1
& Backbone 
& $2.5$K 
& $2.5$--$3.0$ h \\

Phase 2a
& Side heads 
& $2.5$K 
& $2.0$--$2.5$ h \\

Phase 2b 
& Backbone + side heads 
& $12.5$K 
& $16$--$20$ h \\

Phase 3
& Backbone + side heads 
& $2.5$K 
& $6.0$--$7.5$ h \\
\midrule
\textbf{Total} 
& -- 
& $20.0$K 
& $26.5$--$33$ h \\
\bottomrule
\end{tabular}
\end{table}

\section{Details of Reasoning Entropy Estimation}
\label{app:reasoning-entropy}

We provide a detailed definition and implementation of \emph{Reasoning Entropy} in this section. For latent tokens, we consider two possible estimators. The first estimator, \emph{Semantic Region Similarity as Reasoning Entropy}, offers a direct and strict way to quantify how much grounded visual evidence is encoded in each latent token by measuring its compatibility with multiple semantic evidence regions. However, it depends on the semantic decoder \(f_{\mathrm{desc}}\) calibrated for latent states and cannot be directly applied to language tokens due to their next-token prediction nature. Therefore, we use it only as a reference analysis for latent tokens.

The second estimator, \emph{Intervention-based Visual Evidence Importance as Reasoning Entropy}, places latent tokens and language tokens in the same space. It therefore enables a unified computation and comparison of \emph{Reasoning Entropy} across latent tokens, transition tokens, and answer tokens. In practice, we found that the entropy of latent tokens computed by the two estimators exhibits highly consistent trends. Therefore, unless otherwise specified, we adopt the intervention-based estimator as the unified definition of \emph{Reasoning Entropy} throughout our analysis.

\subsection{Dataset Preparation}

For each image--question pair \((I,q)\), we construct an analysis-only bank of
diverse grounded reasoning traces. Following the same data construction protocol
used for our step-wise grounded supervision, we prompt a strong MLLM to sample
multiple visual reasoning traces. Each trace contains a sequence of
grounded evidence steps, and each step consists of a bounding
box, a region-specific semantic description, and the corresponding visual
information. We retain only traces whose final answers match the ground-truth
answer and whose grounded boxes pass the verification procedure. This yields an
image-question-specific trace bank:
\[
\mathcal{T}(I,q)
=
\left\{
\mathcal{T}_m
=
\{(b_{m,s}, d_{m,s}, v_{m,s})\}_{s=1}^{S_m}
\right\}_{m=1}^{M},
\]
where \(m\) indexes a valid reasoning trace, \(s\) indexes an evidence step,
\(b_{m,s}\) is the verified bounding box,
\(d_{m,s}\) is the region description, and \(v_{m,s}\) is the extracted visual
information.

We encode each evidence step into the same semantic space used by the semantic
alignment decoder:
\[
e_{m,s}
=
g\left(d_{m,s} \oplus v_{m,s}\right),
\]
where \(g(\cdot)\) denotes the frozen semantic encoder used to build stable
region-level semantic anchors, and \(\oplus\) denotes textual concatenation.
The resulting set
\[
\mathcal{B}(I,q)=\{e_{m,s}\}_{m=1,s=1}^{M,S_m}
\]
serves as an image-question-specific bank of answer-relevant visual evidence.

\subsection{Semantic Region Similarity as Reasoning Entropy}

\paragraph{Probing latent token states in the semantic space.}
For a latent visual reasoning model, we collect the last-layer hidden states along the
latent trajectory, including both supervised and unsupervised latent tokens. For a hidden state \(h_t\), we use
the calibrated semantic alignment decoder as a probing interface:
\[
d_t = f_{\mathrm{desc}}(h_t).
\]

We compute the similarity between \(d_t\) and each evidence step in
\(\mathcal{B}(I,q)\), and normalize the similarities into an evidence-step
distribution:
\[ \small
p_{m,s}^{(t)}
=
\frac{
\exp\left(\cos(d_t,e_{m,s})/\tau\right)
}{
\sum_{m'=1}^{M}
\sum_{s'=1}^{S_{m'}}
\exp\left(\cos(d_t,e_{m',s'})/\tau\right)
},
\]
where \(\tau\) is a temperature parameter.

\paragraph{Normalized reasoning entropy.}
Our goal is to measure whether a latent state remains compatible with
multiple plausible grounded reasoning traces, we aggregate the evidence-step
distribution into a trace-level distribution:
\[ \small
P_m^{(t)}
=
\sum_{s=1}^{S_m} p_{m,s}^{(t)}.
\]
The reasoning entropy of token \(t\) is then defined as
\[ \small
H_{\mathrm{reason}}(h_t)
=
-
\sum_{m=1}^{M}
P_m^{(t)}
\log P_m^{(t)}.
\]
To make values comparable across samples with different numbers of retained
valid traces, we report the normalized reasoning entropy:
\[ \small
\widetilde{H}_{\mathrm{reason}}(h_t)
=
\frac{
H_{\mathrm{reason}}(h_t)
}{
\log M
}.
\]
Thus, \(\widetilde{H}_{\mathrm{reason}}(h_t)\in[0,1]\). A higher value indicates
that the hidden state remains semantically compatible with multiple valid
grounded reasoning traces, while a lower value indicates that the state has
concentrated on a smaller set of reasoning possibilities.



This probing analysis should not be interpreted as decoding multiple bounding
boxes from one latent token. The visual grounding decoder \(f_{\mathrm{reg}}\)
still predicts a single supervised bounding box for each grounded latent token.
The entropy is instead estimated through the semantic alignment decoder
\(f_{\mathrm{desc}}\), which probes whether the hidden state is semantically
close to multiple valid grounded reasoning traces beyond its explicitly decoded
spatial output.

\subsection{Intervention-based Visual Evidence Importance as Reasoning Entropy}


The semantic probing entropy above measures whether a hidden state is compatible with multiple grounded reasoning traces in the learned semantic evidence space. Although more intuitive, this probing interface relies on the calibrated semantic decoder \(f_{\mathrm{desc}}\), which is trained on latent states and is therefore not directly comparable for ordinary language states. Another possible choice is to compute entropy from the language-modeling distribution, i.e.,
\[
H_{\mathrm{vocab}}(t)
=
-\sum_{v\in\mathcal{V}}
p_{\theta}(v\mid u_t)\log p_{\theta}(v\mid u_t),
\]
where \(p_{\theta}(v\mid u_t)\) is obtained by applying the LM head and softmax to the hidden state. However, this quantity measures next-token prediction uncertainty over the vocabulary rather than the amount of visual evidence represented by the current state. For example, a transition token may have low vocabulary entropy simply because the next word is syntactically predictable, even if its hidden state still depends on multiple visual regions. Conversely, function words or ambiguous lexical continuations may yield high vocabulary entropy without indicating broad visual grounding. Moreover, vocabulary entropy is not naturally defined for non-linguistic latent slots, making it unsuitable for comparing latent tokens and language tokens in a shared representational space.

To fairly compare latent tokens, transition tokens, and answer tokens, we therefore estimate entropy through an intervention-based visual evidence distribution, which does not require any additional projection head or vocabulary-level decoding. This estimate evaluates all analyzed states in the same image-question-specific evidence space by measuring how their representations change when each grounded visual evidence region is removed.

For each image--question pair \((I,q)\), we use the grounded evidence bank \(\mathcal{B}(I,q)\) defined above. Each evidence node corresponds to a verified region \(b_k\) from the retained grounded traces, where \(k\in\{1,\ldots,K_{I,q}\}\). To avoid artificially increasing entropy due to repeated boxes across different traces, we merge highly overlapping evidence regions using non-maximum suppression and keep the merged regions as the intervention units. Let \(u_t\) denote the last-layer hidden state at trajectory position \(t\), where \(u_t\) can be a supervised latent token, an unsupervised latent token, or to generate a transition token or an answer token. The model first inferences on the original image and record the full-state representation \(u_t^{\mathrm{full}}\).

We then construct one counterfactual input for each evidence node by masking the corresponding image region \(b_k\), producing \(I^{(-k)}\). The same question, latent-token layout, transition tokens, and answer tokens are used under teacher forcing, so that token positions are aligned across the original and counterfactual forward passes. This yields a counterfactual representation \(u_t^{(-k)}\) for every analyzed state, and the difference between \(u_t^{\mathrm{full}}\) and \(u_t^{(-k)}\) reflects the effect of removing visual evidence region \(b_k\), rather than changes in the generated token sequence. The visual sensitivity of state \(u_t\) to evidence node \(k\) is defined as
\[
s_{t,k}
=
\max\left(
0,\,
1-\cos\left(
\mathrm{LN}(u_t^{\mathrm{full}}),
\mathrm{LN}(u_t^{(-k)})
\right)
\right),
\]
where \(\mathrm{LN}(\cdot)\) denotes the same final hidden-state normalization used before the language modeling head. A larger \(s_{t,k}\) indicates that removing region \(b_k\) induces a larger change in the token state, suggesting that the state depends more strongly on this visual evidence node.

We normalize the sensitivities over all evidence nodes to obtain an interventional visual evidence distribution:
\[
q_{t,k}
=
\frac{s_{t,k}+\epsilon}
{\sum_{r=1}^{K_{I,q}}(s_{t,r}+\epsilon)} ,
\]
where \(\epsilon\) is a small constant for numerical stability. The corresponding evidence entropy is
\[ \small
H_{\mathrm{IVE}}(u_t)
=
-\sum_{k=1}^{K_{I,q}}
q_{t,k}\log q_{t,k}.
\]
Since different samples may contain different numbers of retained evidence nodes, we report normalized entropy:
\[ \small
\widetilde{H}_{\mathrm{IVE}}(u_t)
=
\frac{H_{\mathrm{IVE}}(u_t)}
{\log K_{I,q}} .
\]

A potential issue is that visually inactive tokens, such as function words or punctuation, may have uniformly small sensitivities to all evidence regions. Such tokens can obtain spuriously high normalized entropy after normalization. We therefore compute the total visual sensitivity mass
\[ \small
M_t=\sum_{k=1}^{K_{I,q}}s_{t,k},
\]
and use a mass-aware entropy score:
\[ \small
\widehat{H}_{\mathrm{IVE}}(u_t)
=
\frac{M_t}{M_t+\alpha}
\cdot
\widetilde{H}_{\mathrm{IVE}}(u_t),
\]
where \(\alpha\) is set to the median visual sensitivity mass over all analyzed states on the validation subset. This weighting preserves high entropy only when the token state is both visually grounded and broadly influenced by multiple evidence nodes.

\paragraph{Trajectory-level aggregation.}
We compute \(\widehat{H}_{\mathrm{IVE}}\) for all states along the latent-to-answer trajectory. Fixed latent slots are averaged by slot index across samples. Since the number of transition and answer tokens may vary across examples, we align textual positions by their normalized phase position and average them into fixed-width bins. Special tokens and padding tokens are excluded. The plotted curve reports the mean entropy over samples, with the latent phase, transition-token phase, and answer-token phase shown in order.

This intervention-based estimate places latent tokens and language tokens in the same evidence space: both are evaluated by how their hidden states causally respond to removing each answer-relevant visual region. Therefore, the entropy gap should be interpreted as evidence-dispersion rather than next-token uncertainty. High values indicate that a state remains sensitive to multiple grounded visual evidence nodes, while low values indicate that the state has concentrated on a narrower set of evidence required for final vocabulary-aligned decoding.

\section{Dataset Statistics and Visualization}
\label{dataset_viz}

\begin{figure}
    \centering
    \includegraphics[width=0.8\linewidth]{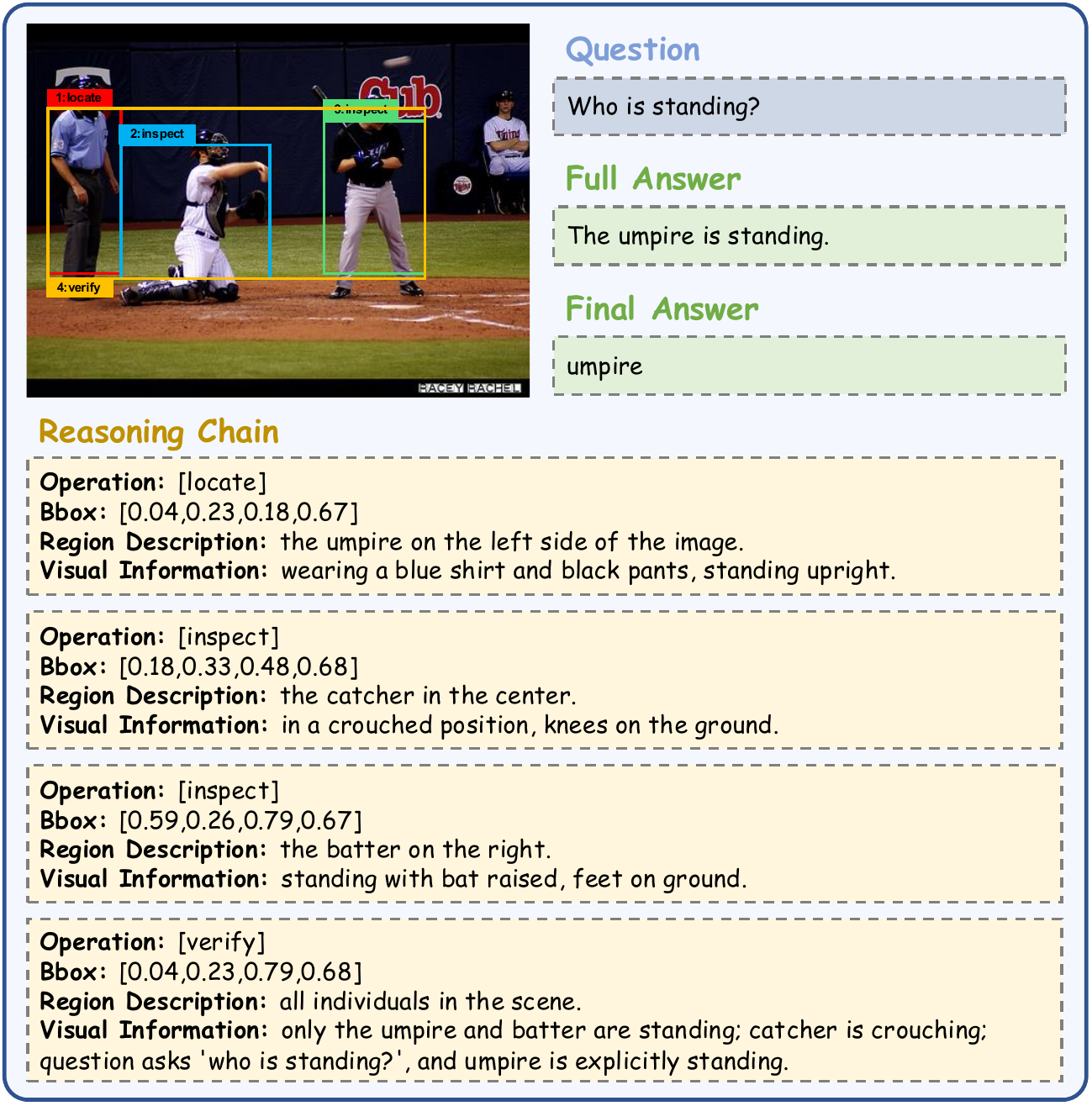}
    \caption{Example of 4-steps Supervision Sample.}
    \label{fig:dataset-viz-1}
\end{figure}

\begin{figure}
    \centering
    \includegraphics[width=0.8\linewidth]{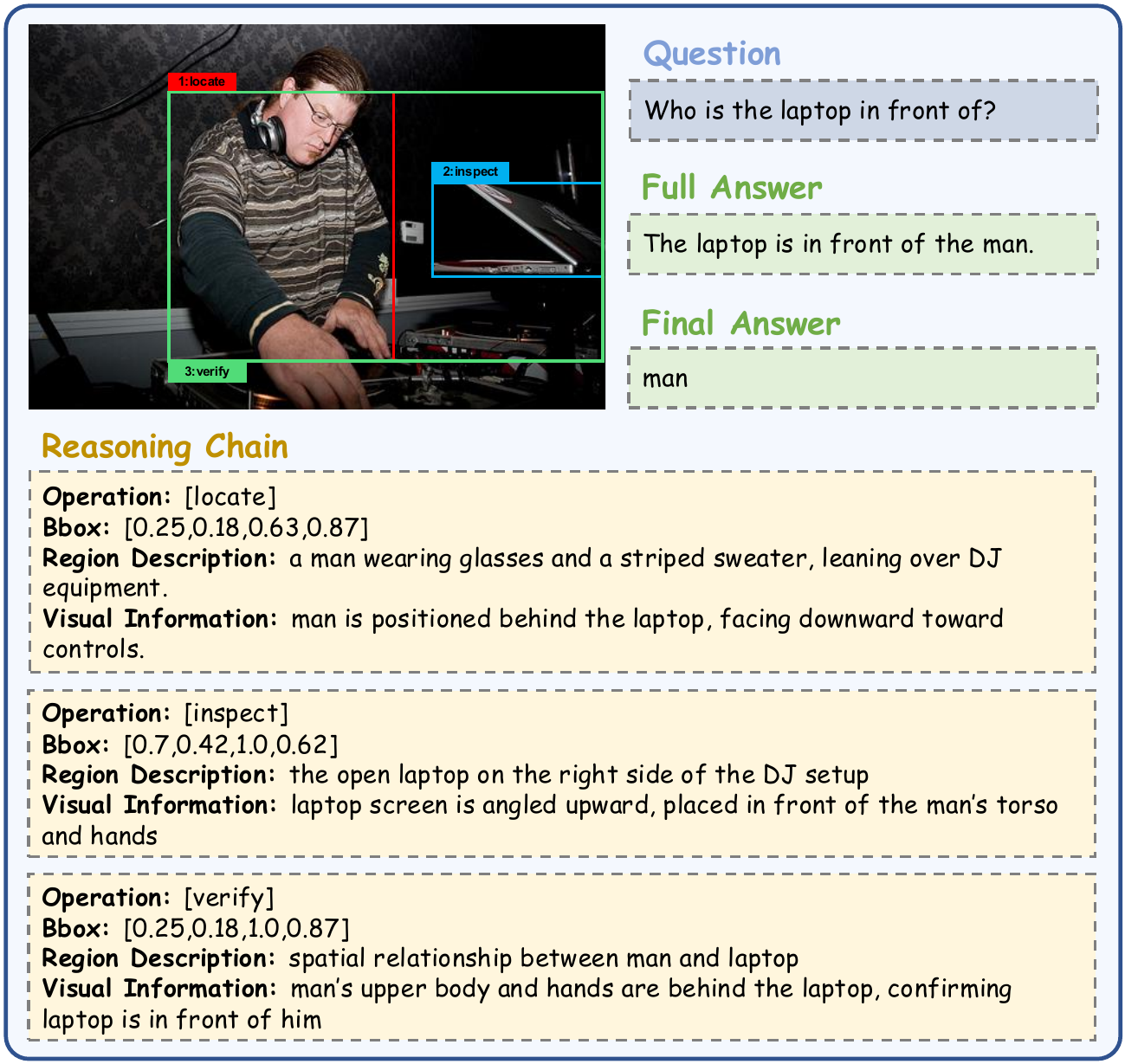}
    \caption{Example of 3-steps Supervision Sample.}
    \label{fig:dataset-viz-2}
\end{figure}

\begin{figure}
    \centering
    \includegraphics[width=0.8\linewidth]{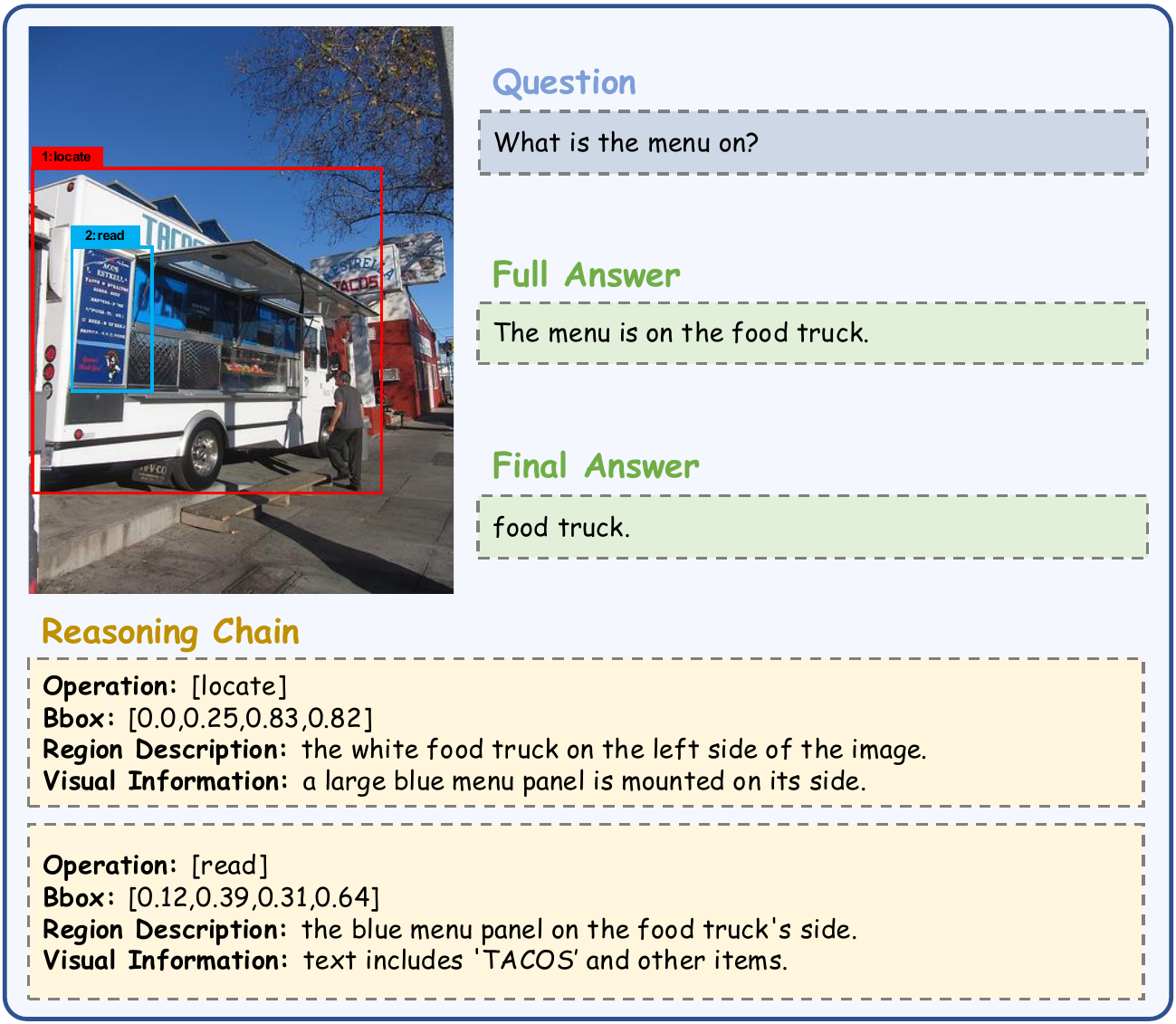}
    \caption{Example of 2-steps Supervision Sample.}
    \label{fig:dataset-viz-3}
\end{figure}

Tables~\ref{tab:glsd-statistics}--\ref{tab:glsd-step-fields} summarize the overall scale, reasoning-step distribution, JSONL sample schema, and per-step annotation format of GLSD. Figure~\ref{fig:dataset-viz-1}--\ref{fig:dataset-viz-3} provide visualization of GLSD.

\begin{table}[t]
\centering
\small
\caption{Statistics of the \textit{Grounded Latent Supervision Dataset} (GLSD).}
\label{tab:glsd-statistics}
\begin{tabular}{lc}
\toprule
Item & Value \\
\midrule
Source dataset & GQA train split \\
Storage format & JSONL \\
Parseable samples & 96,000 \\
Reasoning steps per sample & 2--4 \\
Total grounded reasoning steps & 256,444 \\
Average reasoning steps & 2.67 \\
Spatial supervision & Normalized and pixel-level bounding boxes \\
Semantic supervision & Region descriptions and visual information \\
Answer fields &  full answer and final answer \\
\bottomrule
\end{tabular}
\end{table}

\begin{table}[t]
\centering
\small
\caption{Distribution of reasoning-chain lengths in GLSD.}
\label{tab:glsd-step-distribution}
\begin{tabular}{lcc}
\toprule
Reasoning-chain length & Number of samples & Percentage \\
\midrule
2 steps & 47,410 & 49.39\% \\
3 steps & 32,736 & 34.10\% \\
4 steps & 15,854 & 16.51\% \\
\midrule
Total & 96,000 & 100.00\% \\
\bottomrule
\end{tabular}
\end{table}

\begin{table}[t]
\centering
\small
\caption{Top-level fields in each GLSD JSONL sample.}
\label{tab:glsd-top-fields}
\begin{tabular}{lll}
\toprule
Field & Type & Description \\
\midrule
\texttt{question} & string & Question text \\
\texttt{answer} & string & final answer phrase \\
\texttt{full\_answer} & string & elaborated answer (transition) \\
\texttt{image} & string & GQA image filename \\
\texttt{width}, \texttt{height} & int & Image resolution \\
\texttt{dataset}, \texttt{split} & string & Source dataset and split \\
\texttt{reasoning\_chain} & array & Step-wise grounded reasoning trace \\
\texttt{annotation\_mask} & array[int] & Valid-step mask padded to budget $K$ \\
\texttt{K} & int & Latent-slot budget in the stored sample \\
\texttt{reasoning\_chain\_viz\_file} & string & Optional visualization filename \\
\bottomrule
\end{tabular}
\end{table}

\begin{table}[t]
\centering
\small
\caption{Fields of each reasoning step in \texttt{reasoning\_chain}.}
\label{tab:glsd-step-fields}
\begin{tabular}{lll}
\toprule
Field & Type & Description \\
\midrule
\texttt{step} & int & Step index starting from 1 \\
\texttt{operation} & string & Step type, e.g., \texttt{locate}, \texttt{inspect}, \texttt{verify} \\
\texttt{bbox\_01} & array[float] & Normalized box $[x_1,y_1,x_2,y_2]$ in $[0,1]$ \\
\texttt{bbox\_pixels} & array[int] & Pixel-space bounding box \\
\texttt{region\_description} & string & Description of the attended region \\
\texttt{visual\_information} & string & Visual evidence extracted from the region \\
\bottomrule
\end{tabular}
\end{table}

\end{document}